\documentclass[a4paper]{article}

\usepackage{INTERSPEECH_v2}

\usepackage[utf8]{inputenc}
\usepackage[hidelinks]{hyperref}
\usepackage[none]{hyphenat} 

\title{Articulation rate in Swedish child-directed speech increases as a function of the age of the child even when surprisal is controlled for\thanks{$^*$This work was carried out while this author was visiting the Department of Linguistics, Stockholm University.}}

\name{Johan Sjons$^1$, Thomas Hörberg$^1$, Robert Östling$^1$ and Johannes Bjerva$^{2*}$}
\address{
  $^1$Department of Linguistics, Stockholm University, Sweden\\
  $^2$Center for Language and Cognition Groningen, University of Groningen, The Netherlands}
\email{\{johan|thomas\_h|robert\}@ling.su.se,j.bjerva@rug.nl}

\begin{document}

\maketitle
\begin{abstract}
In earlier work, we have shown that articulation rate in Swedish child-directed speech (CDS) increases as a function of the age of the child, even when utterance length and differences in articulation rate between subjects are controlled for. In this paper we show on utterance level in spontaneous Swedish speech that i) for the youngest children, articulation rate in CDS is lower than in adult-directed speech (ADS), ii) there is a significant negative correlation between articulation rate and surprisal (the negative log probability) in ADS, and iii) the increase in articulation rate in Swedish CDS as a function of the age of the child holds, even when surprisal along with utterance length and differences in articulation rate between speakers are controlled for. These results indicate that adults adjust their articulation rate to make it fit the linguistic capacity of the child.
\end{abstract}

\section{Introduction}
When speaking, we make choices about how to package whatever message we are about to send. One principled way of predicting how we make these choices is to apply the Hyper- and Hypospeech (H\&H) theory. 
This theory predicts that speakers continuously approximate how much the speech signal must be adjusted in order for it to fit the recipient's capacity to decode it, and that speakers do adjust their linguistic signal to fit this approximation accordingly, along a continuum of hyper- and hypospeech \cite{lindblom1990explaining}. One certain type of recipient is the pre-linguistic child, who has a lower capacity to decode the linguistic signal than older children and adults. One way of measuring the degree of such an adjustment within an utterance is \emph{articulation rate} (AR). 

In this longitudinal corpus-based study, we investigate whether adults increase their AR in Swedish child-directed speech (CDS), as a function of the linguistic capacity of the child, approximated to age, for children aged 0;7--2;9, controlling for factors influencing AR.

\section{Related work}
It has long been recognized that CDS has several distinct characteristics compared to adult-directed speech (ADS), such as fewer words per utterance \cite{phillips1973syntax} and lower speech rate \cite{broen1972verbal}. Furthermore, AR has been shown to be lower in ADS than in CDS, \cite{fernald1984expanded, narayan2016speech, van1997language}, which is a measure of speech tempo, defined as the number of linguistic units per time unit, excluding pauses. Speech rate is the number of linguistic units per time unit, including pauses (e.g., \cite{goldman1961significance, trouvain2001articulation}).


However, less is known about the extent to which the characteristics of CDS change as the child grows older. In the case of AR, it has been shown that it increases in mothers' (n=16) CDS as a function of the age of the child for children from 0;4 to 1;4 of age in Korean (n=6), Sri Lankan Tamil (n=5) and Tagalog (n=5). This is referred to as ``speech rate'', but it is stated in a footnote that what is actually meant is AR, under the constraint of including utterances where silences never exceeded a duration of 300 ms. Utterance length was controlled for by choosing utterances around 5 s long \cite{narayan2016speech}.

In a corpus-based longitudinal study on AR in Swedish CDS \cite{sjons2016articulation}, it was shown that AR increased as a function of the age of the child, even when utterance length in terms of number of syllables was controlled for, something which has been shown to influence AR, in the sense that longer utterances in terms of number of linguistic units tend to have higher AR than shorter ones \cite{lindblom1973some,quene2008multilevel}. If utterances contained pauses, these were never of duration \textgreater 200 ms. Differences in AR between speakers were also controlled for, since AR between speakers is known to vary \cite{hilton2011syllable}. Children were between 0;7 and 2;9 of age. 

Hence, there are at least two factors that can influence AR: Utterance length in terms of linguistic units, and individual AR. However, given the H\&H theory's prediction that speakers adapt the speech signal according to their estimation of the recipient's capacity to decode it, it is reasonable to believe that speakers also adjust their AR in accordance with how surprising what they are about to say is. Intuitively, if what they are about to say is less surprising, they would typically use a higher AR, and vice versa, assuming an overall \emph{information rate} which is achieved by balancing AR and how surprising the utterance is \cite{karlgren1961speech,pellegrino2011across}, the latter which as been referred to as \emph{surprisal} \cite{levy2007speakers}. Surprisal can be specified in information-theoretic terms, that is, as the negative log of the probability of a word \emph{n}, given the probability of the preceding word(s) \cite{levy2007speakers}. This does in essence go back to Shannon's influential theory of communication \cite{shannon1948mathematical}:

\begin{equation}
  -\log_2 P(w_n|w_1 ... w_{n-1})
  \label{eq1}
\end{equation}

\noindent In the present study, we estimate surprisal with a language model (see Section \ref{languagemodeling}).


\subsection{Research Questions}
We ask three questions: 1) In adult-directed speech, is there a correlation between articulation rate and utterance length, on the one hand, and articulation rate and surprisal, on the other? 2) Is articulation rate in adult-directed speech higher than in child-directed speech, when controlling for utterance length, differences in articulation rate between speakers and surprisal? 3) Does the increase in articulation rate in child-directed speech as a function of child age shown in \cite{sjons2016articulation} hold, even when controlling for surprisal?

\section{Method and Data}
To answer these questions, we needed Swedish data on AR in ADS and CDS respectively, and a model of how surprisal is represented in the speaker-hearer. We built two different models that were trained and tested. Hence, the data fed to the language models in this study consist of five different sets. Three of these form the training set for the models (see Section \ref{sec:training}). The other two form the test sets (see Section \ref{sec:testing}), which both are parts of two different corpora. One, henceforth \emph{the CDS test set}, is from the MINGLE-corpus \cite{nilsson2014multimodal}. The other is half of Spontal \cite{edlund2010spontal}, henceforth \emph{the ADS test set}.

\subsection{Corpora}
The MINGLE-corpus is a corpus that has been multi-modally annotated from recordings collected at Stockholm Babylab which is part of the Phonetics Laboratory at Stockholm University. Recordings consist of parent-child dyads in which the subjects were equipped with wireless microphones which let them move around freely. The sessions between parent and child consisted of some task, such as doing a jigsaw-puzzle and of free play. The sessions lasted for about half an hour \cite{nilsson2014multimodal, lacerda2009}.


Spontal is a corpus consisting of transcribed face-to-face dialogues between two speakers at a time, in which the maximum length of a silence within an utterance is 200 ms \cite{edlund2010spontal, edlundPersonal}. Utterances from 12 of the 24 of these dialogues were used in the ADS test set. Utterances containing laughter, and coughing were excluded.

Swedish Blog Sentences (SBS) is a corpus containing \texttildelow2.7 billion tokens, from randomly rearranged sentences \cite{ostling2013compounding}.

\subsection{Language model training data}
\label{sec:training}
We train the language models on three data sets: Half of a subset of Spontal (the other subset we test on), a subset of material similar to the MINGLE-corpus, and the first 1,400,000 lines of SBS \cite{ostling2013compounding} (see Table~\ref{tab:langmod}).

\begin{table}[h]
  \caption{Training data used for language modeling.}
  \label{tab:langmod}
  \centering
  \begin{tabular}{llr}
    \toprule
    \textbf{Corpus} & \textbf{Description} & \textbf{N tokens used} \\
    \midrule
    Spontal                   & ADS & 13,300             \\
    CDS                    & CDS & 6,063       \\
    Blogs                     & Social media text  & 16,188,877               \\
    \midrule
    Total                     &   & 16,208,240 \\
    \bottomrule
  \end{tabular}
\end{table}

\subsection{Language model test data}
\label{sec:testing}
Transcribed utterances from the MINGLE-corpus of 7 parents, 4 mothers and 3 fathers (i.e., 4 children), in a total of 28 sessions, were modified to a form more faithful to spoken Swedish (e.g., \emph{någonting} (something) was changed to \emph{nåt} in relevant cases). Also, utterances were shortened, prolonged or split so as not to contain pauses of duration longer than 200 ms. Utterances containing laughter, onomatopoetic sounds, singing and whispers were omitted. Only utterances from when the parent and child were alone in the room were included, so as to avoid speech from when the parent communicated with the experimental leader through the child.

The number of syllables for both test sets were approximated as the number of transcribed vowels in each utterance, which in Swedish gives a very good approximation, since in principle every orthographic vowel corresponds to a realized vowel. Articulation rate was defined as the number of transcribed vowels per utterance over the length in milliseconds for that utterance. For each session, a script extracted the age of the child in days, AR for every utterance, and the number of syllables per utterance. This resulted in 7,865 utterances containing in total 29,000 tokens for the CDS test set, and 3,087 utterances containing 16,301 tokens in total for the ADS test set.


\subsection{Language Modeling}
\label{languagemodeling}
We applied a character-level neural language model.
The model employs a convolutional neural network (CNN) and a highway network over characters, whose output is given to a long short-term memory (LSTM) recurrent neural network language model \cite{kim:2016}.
The model takes a sequence of characters as input, and outputs the probability of that sequence.\footnote{We use the implementation found at \url{https://github.com/jarfo/kchar}.}
One reason for opting for a character-level model is that words which are not seen in the training material can still be modeled well.
For instance, if the model has seen words such as \textit{quick}, \textit{sick}, and \textit{quickly}, it would be able to assign a well-estimated probability to the unseen word \textit{sickly}.

For purposes of comparison, we also applied a language model which uses word-level trigram statistics with Kneser-Ney smoothing \cite{KneserNey1995}, using the formulation of \cite{ChenGoodman1999}.







\section{Results and Discussion}
All analyses in this section were performed twice, using the surprisal estimates of both the Neural Character-Based Language Model and the Kneser-Ney N-gram Language Model. Due to space limitations we only report analyses using the former. Unless explicitly stated, the analyses using the latter did not qualitatively differ from the former. Since most of the training data of the language models consisted of written blog data or adult-directed speech, there is reason to assume that the models provide more accurate estimates of surprisal for the ADS test set, than for the CDS test set. If the relationship between AR in CDS and our estimates of surprisal is similar to the relationship between AR in ADS and surprisal, however, we can be more confident that the surprisal estimates are applicable also to CDS test set. We therefore present an analysis of the ADS test set first, then move on to the analysis of the CDS test set, and finally look at the difference between ADS and CDS.

\subsection{ADS test set}
\begin{figure}[h!]
	\centering
	\includegraphics[width=0.9\linewidth]{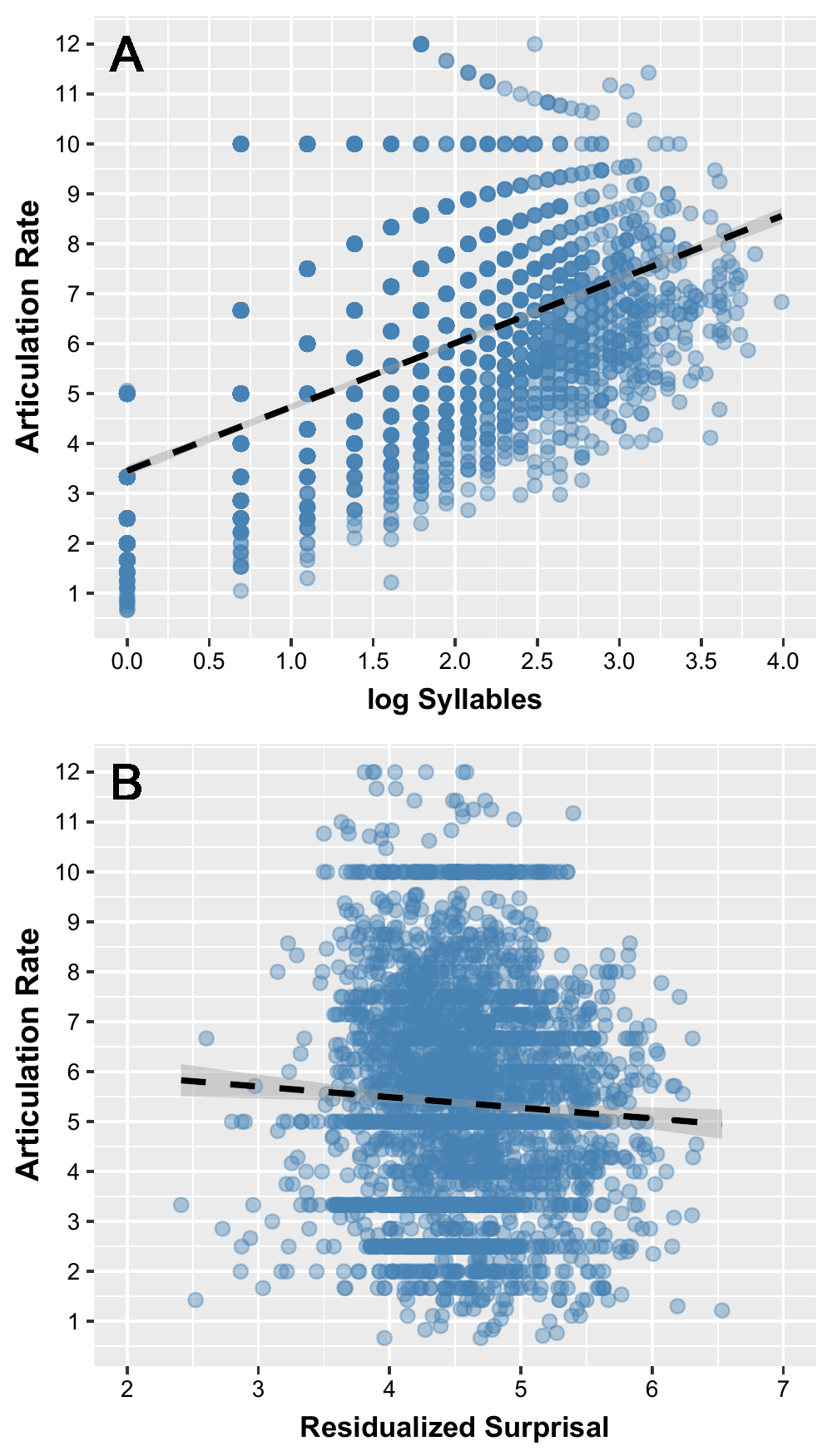}
	\caption{Articulation rate in ADS data from the Spontal corpus as a function of utterance length in terms of log syllables (Panel A), and Surprisal corrected for utterance length (Panel B).}
	\label{fig:AdultRate}
\end{figure}

Figure \ref{fig:AdultRate} illustrates the relationship between AR and utterance length in terms of log syllables (Panel A), on the one hand, and AR and surprisal corrected for utterance length (Panel B), on the other, in the ADS test set.

Figure \ref{fig:AdultRate} indicates that, in ADS speech, AR is overall higher for longer utterances, but, on the other hand, lower for utterances with high surprisal. These relationships were evaluated with mixed effects modeling \cite{GelmanHill2006}. The mixed effects model is a type of general linear model \cite{Howell2010} that accounts for random effects, such as differences in AR between speakers. It is therefore possible to rule out that any observed relationships do not stem from random speaker differences (e.g., that one or a few speakers with a high AR also tend to use long utterances in the present data).
The model predicts AR in ADS as a function of utterance length in terms of standardized log syllables \textsc{(stdLogSyllables)}, surprisal \textsc{(Surprisal)} and their interaction \textsc{(StdLogSyllables:Surprisal)}. The model also included a random intercept for speaker and by-speaker random slopes for \textsc{stdLogSyllables} and \textsc{Surprisal}.\footnote{The random effects structure was determined on the basis of backward elimination of random effects. Only those effects that significantly improved the model's predictive ability were included.} In other words, the model controls for AR differences between speakers, as well as speaker differences in the influence of utterance length and surprisal on AR.
Analyses were conducted in the statistical language R (v. 3.1.2) \cite{r2014}. Degrees of freedom for the calculation of p-values were estimated using Welch-Satterthwaite approximation, as implemented in the lmerTest() package (v. 2.0-20) \cite{kuznetsova2014}. \textsc{Articulation Rate} and \textsc{Surprisal} that deviated by more than 3 standard deviations from their overall means were excluded.
The model found a significant effect of \textsc{stdlogSyllables}, $\beta = 3.68, t(1488.4) = 14.64, p < .0001$, a significant effect of \textsc{Surprisal}, $\beta = -0.3, t(17) = -3.37, p < .001$, and a significant \textsc{stdLogSyllables}:\textsc{Surprisal} interaction $\beta = -0.51, t(2861.5) = -9.32, p <.0001$. 
These results thus confirm the findings of \cite{sjons2016articulation} for adult directed speech by showing that longer utterances are on average articulated faster than shorter utterances. They also show that utterances that are less expected in terms of \textsc{Surprisal} are articulated somewhat slower, and this is in particular the case for longer utterances (as shown by the \textsc{stdLogSyllables}:\textsc{Surprisal} interaction). This is to be expected and in line with the findings of \cite{pellegrino2011across}.

\subsection{CDS test set}
\begin{figure}[h!]
	\centering
	\includegraphics[width=0.9\linewidth]{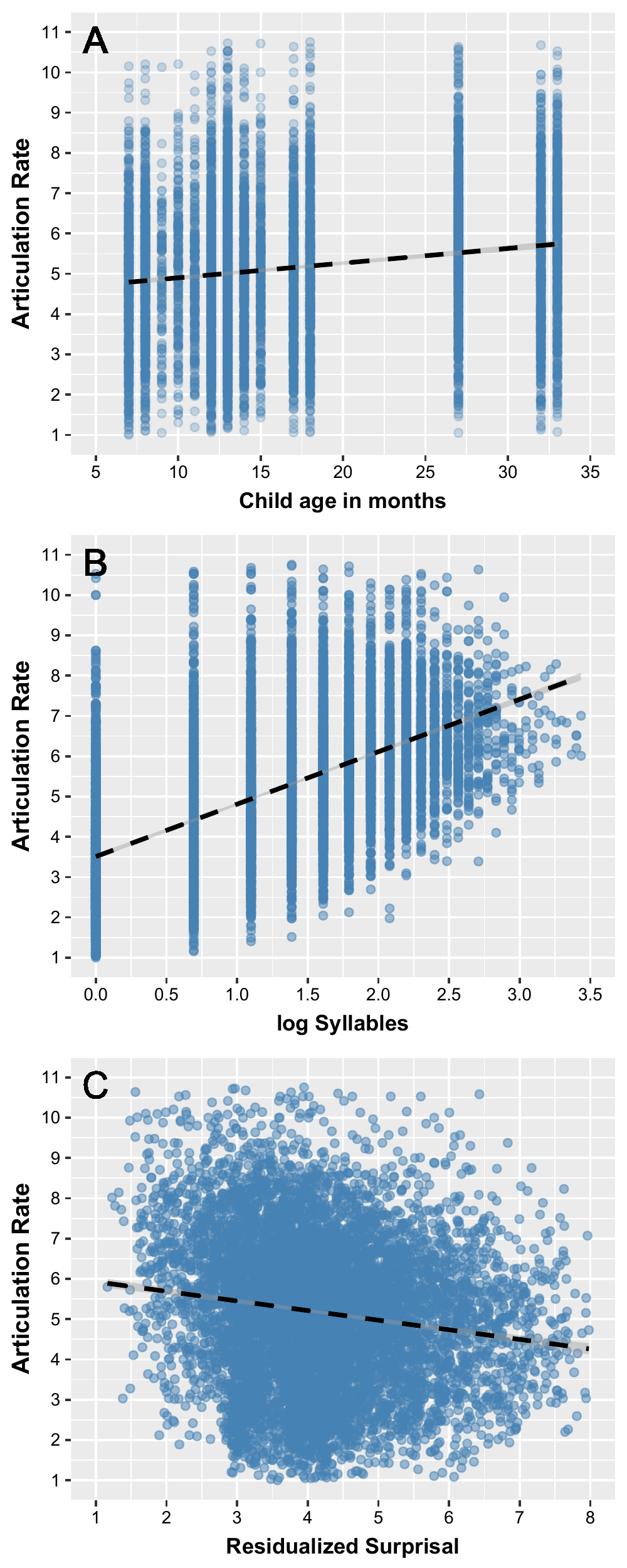}
	\caption{Articulation rate in the CDS test set as a function of child age in months (Panel A), utterance length in terms of log syllables (Panel B), and Surprisal corrected for utterance length (Panel C).}
	\label{fig:ChildRate}
\end{figure}

Figure \ref{fig:ChildRate} shows the relationship between AR and child age in months (Panel A), AR and utterance length in terms of log syllables (Panel B), and, finally, AR and surprisal corrected for utterance length (Panel C) in the CDS test set.

Panel A and B show that, as found in \cite{sjons2016articulation}, there is a slight increase in AR in CDS as the child gets older, and that, as in ADS, AR is higher for longer utterances. Panel C indicates that also in CDS, AR is lower for utterances that are less expected in terms of surprisal. 
In order to determine whether the relationship between child age and AR in CDS still holds when also surprisal is controlled for, the CDS test set was analyzed with mixed effects modeling. This model predicts AR in CDS as a function of utterance length, again in terms of standardized log syllables (\textsc{stdLogSyllables}), surprisal (\textsc{Surprisal}), and child age (\textsc{Age}) and the interactions of these factors. The model also contains a random speaker intercept and by-speaker random slopes for \textsc{stdLogSyllables} and the \textsc{StdLogSyllables}:\textsc{Age} interaction. Again, \textsc{Articulation Rate} and \textsc{Surprisal} that deviated by more than 3 standard deviations from their overall means were excluded.
This model also found a significant effect of \textsc{stdlogSyllables}, $\beta = 1.8, t(16) = 19.902, p < .0001$, a significant effect of \textsc{Surprisal}, $\beta = -0.23, t(1956) = -16, p < .0001$, and a significant \textsc{stdLogSyllables}:\textsc{Surprisal} interaction, $\beta = -0.15, t(7635) = -10.6, p <.0001$. In other words, the relationship between AR, utterance length and Surprisal in CDS is more or less the same as in ADS: Longer utterances are on the average articulated faster than shorter ones. Utterances that are less expected in terms of \textsc{Surprisal} are articulated somewhat slower, and this is in particular the case for longer utterances. Importantly, the model also found a significant effect of Age, $\beta = 0.31, t(1707) = 4.67, p < .05$, and a significant \textsc{Age}:\textsc{Surprisal} interaction, $\beta = -0.07, t(9) = -2.98, p <.05$ (showing that the increase in AR with child age is lower for surprising utterances). Importantly, the 
relationship between child age and utterance length still holds when surprisal is controlled for. That is, the age effect is not an artifact of utterances directed towards older children on average being somewhat more expected, and consequently articulated somewhat faster.

\subsection{Articulation rate in CDS in comparison to ADS}
Finally, we investigate if AR is lower in CDS than in ADS. As shown in the previous sections, AR is in both ADS and CDS much influenced by utterance length, on the one hand, and surprisal, on the other. In order to make a comparison between AR in CDS and ADS, these factors need to be controlled for. Using a combination of the CDS test set and the ADS test set, we therefore fitted a model that predicts AR on the basis of speech type (CDS vs. ADS), utterance length in standardized log syllables \textsc{stdlogSyllables}, \textsc{Surprisal} and the \textsc{stdlogSyllables}:\textsc{Surprisal} interaction. The model also included random speaker intercepts and by-speaker random slopes for \textsc{stdLogSyllables}, \textsc{Surprisal} and their interaction. As expected, the model found a significant effect of \textsc{stdlogSyllables}, $\beta = 3.02, t(14.94) = 11.69, p < .0001$, a significant effect of \textsc{Surprisal}, $\beta = -0.27, t(5.99) = -6.4, p < .001$, and a significant \textsc{stdLogSyllables}:\textsc{Surprisal} interaction, $\beta = -0.37, t(11.05) = -7.18, p <.0001$, confirming the results of analyses conducted across each data set individually. Importantly, the model also showed a significant effect of \textsc{Speechtype} (CDS being the reference category), $\beta = -0.56, t(20.29) = -4.58, p <.001$, showing that ADS is articulated significantly faster than CDS even when utterance length, \textsc{Surprisal} and their interaction are controlled for. It should be noted, however, that the model conducted using the \textsc{Surprisal} estimates from the Kneser-Ney N-gram Language Model failed to find a significant effect of speech type. This might very well be due to the model's inability to determine accurate surprisal estimates for short and highly infrequent utterances, as indicated by the astonishingly high surprisal estimates for some of the utterances (with surprisal values as high as 30 bits in some cases). 

\subsection{Drawbacks and Future Work}
A limitation of the language models is that they only model subsequent probabilities based on linearly preceding linguistic output without consideration of syntactic structure or pragmatic constraints.

Although we assumed the function mapping AR to age to be linear, this is not necessarily the case. There could very well be local increases and decreases in AR over time.

Future work would benefit from taking into account data from what the child produces or meta-data about its linguistic capacity, since there of course is no one-to-one mapping between age and linguistic capacity.

\section{Conclusions}
We have shown that in adult-directed speech, there is a positive correlation between articulation rate and utterance length (in terms of number of linguistic units), and a negative correlation between articulation rate and surprisal. Also, we have shown that articulation rate in adult-directed speech is higher than in child-directed speech, when controlling for utterance length, differences in articulation rate between speakers, and surprisal. Furthermore, this is the first study to show that articulation rate increases as a function of child age, when above-mentioned factors are controlled for. These results lend support to the H\&H theory, since they indicate that speakers adjust their speech signal to make it fit the recipients' approximated capacity to decode the message sent.

\section{Acknowledgments}
Thanks to Mats Wir{\'e}n, Anders Eriksson, Ellen Marklund, El\'{i}sabet Eir Cortes, Calle B{\"o}rstell and anonymous reviewers for helpful comments, and to Mats Wir{\'e}n, Kristina Nilsson Bj{\"o}rkenstam, Stockholm Babylab and KTH Speech, Music and Hearing for sharing data. All remaining errors are our own.



\bibliographystyle{IEEEtran}

\bibliography{mybib}

\begin{thebibliography}{10}
\providecommand{\url}[1]{#1}
\csname url@samestyle\endcsname
\providecommand{\newblock}{\relax}
\providecommand{\bibinfo}[2]{#2}
\providecommand{\BIBentrySTDinterwordspacing}{\spaceskip=0pt\relax}
\providecommand{\BIBentryALTinterwordstretchfactor}{4}
\providecommand{\BIBentryALTinterwordspacing}{\spaceskip=\fontdimen2\font plus
\BIBentryALTinterwordstretchfactor\fontdimen3\font minus
  \fontdimen4\font\relax}
\providecommand{\BIBforeignlanguage}[2]{{%
\expandafter\ifx\csname l@#1\endcsname\relax
\typeout{** WARNING: IEEEtran.bst: No hyphenation pattern has been}%
\typeout{** loaded for the language `#1'. Using the pattern for}%
\typeout{** the default language instead.}%
\else
\language=\csname l@#1\endcsname
\fi
#2}}
\providecommand{\BIBdecl}{\relax}
\BIBdecl

\bibitem{lindblom1990explaining}
B.~Lindblom, ``{Explaining phonetic variation: A sketch of the H\&H theory},''
  in \emph{Speech production and speech modelling}.\hskip 1em plus 0.5em minus
  0.4em\relax Springer, 1990, pp. 403--439.

\bibitem{phillips1973syntax}
J.~R. Phillips, ``{Syntax and vocabulary of mothers' speech to young children:
  Age and sex comparisons},'' \emph{Child development}, pp. 182--185, 1973.

\bibitem{broen1972verbal}
P.~A. Broen, ``{The Verbal Environment of the Language-Learning Child},''
  \emph{ASHA Monographs}, vol.~17, 1972.

\bibitem{fernald1984expanded}
A.~Fernald and T.~Simon, ``{Expanded intonation contours in mothers' speech to
  newborns},'' \emph{Developmental psychology}, vol.~20, no.~1, p. 104, 1984.

\bibitem{narayan2016speech}
C.~R. Narayan and L.~C. McDermott, ``Speech rate and pitch characteristics of
  infant-directed speech: Longitudinal and cross-linguistic observations,''
  \emph{The Journal of the Acoustical Society of America}, vol. 139, no.~3, pp.
  1272--1281, 2016.

\bibitem{van1997language}
J.~Van~de Weijer, ``{Language input to a prelingual infant},'' in \emph{the
  GALA'97 Conference on Language Acquisition}.\hskip 1em plus 0.5em minus
  0.4em\relax Edinburgh University Press, 1997, pp. 290--293.

\bibitem{goldman1961significance}
F.~Goldman-Eisler, ``{The significance of changes in the rate of
  articulation},'' \emph{Language and Speech}, vol.~4, no.~3, pp. 171--174,
  1961.

\bibitem{trouvain2001articulation}
J.~Trouvain, J.~Koreman, A.~Erriquez, and B.~Braun, ``{Articulation rate
  measures and their relation to phone classification in spontaneous and read
  german speech},'' in \emph{ISCA Tutorial and Research Workshop (ITRW) on
  Adaptation Methods for Speech Recognition}, 2001.

\bibitem{sjons2016articulation}
J.~Sjons and T.~H{\"o}rberg, ``Articulation rate in child-directed speech
  increases as a function of child age,'' in \emph{Fonetik 2016}, 2016.

\bibitem{lindblom1973some}
B.~Lindblom and K.~Rapp, ``Some temporal regularities of spoken swedish,''
  \emph{Auditory Analysis and Speech Perception (London, 1975)}, pp. 387--96,
  1973.

\bibitem{quene2008multilevel}
H.~Quen{\'e}, ``Multilevel modeling of between-speaker and within-speaker
  variation in spontaneous speech tempo,'' \emph{The Journal of the Acoustical
  Society of America}, vol. 123, no.~2, pp. 1104--1113, 2008.

\bibitem{hilton2011syllable}
N.~H. Hilton, A.~Sch{\"u}ppert, and C.~Gooskens, ``Syllable reduction and
  articulation rates in danish, norwegian and swedish,'' \emph{Nordic Journal
  of Linguistics}, vol.~34, no.~02, pp. 215--237, 2011.

\bibitem{karlgren1961speech}
H.~Karlgren, ``Speech rate and information theory.''\hskip 1em plus 0.5em minus
  0.4em\relax Proceedings of the 4th International Congress of Phonetic
  Sciences (ICPhS), 1961, pp. 671--77.

\bibitem{pellegrino2011across}
F.~Pellegrino, C.~Coup{\'e}, and E.~Marsico, ``A cross-language perspective on
  speech information rate,'' \emph{Language}, vol.~87, no.~3, pp. 539--558,
  2011.

\bibitem{levy2007speakers}
R.~Levy and T.~F. Jaeger, ``Speakers optimize information density through
  syntactic reduction,'' \emph{Advances in neural information processing
  systems}, vol.~19, p. 849, 2007.

\bibitem{shannon1948mathematical}
C.~E. Shannon, ``A mathematical theory of communication,'' \emph{Bell Systems
  Technical Journal}, vol.~27, pp. 623--656, 1948.

\bibitem{nilsson2014multimodal}
K.~Nilsson~Bj{\"o}rkenstam and M.~Wir{\'e}n, ``{Multimodal Annotation of
  Synchrony in Longitudinal Parent--Child Interaction},'' in \emph{MMC 2014
  Multimodal Corpora: Combining applied and basic research targets: Workshop at
  LREC 2014}, 2014.

\bibitem{edlund2010spontal}
J.~Edlund, J.~Beskow, K.~Elenius, K.~Hellmer, S.~Str{\"o}mbergsson, and
  D.~House, ``Spontal: A swedish spontaneous dialogue corpus of audio, video
  and motion capture.'' in \emph{LREC}, 2010, pp. 2992--2995.

\bibitem{lacerda2009}
F.~Lacerda, ``{On the emergence of early linguistic functions : A biologic and
  interactional perspective},'' \emph{Brain Talk: : Discourse with and in the
  brain}, vol. s. 207-230, 2009.

\bibitem{edlundPersonal}
J.~Edlund, personal communication, 2014-10-21.

\bibitem{ostling2013compounding}
R.~{\"O}stling and M.~Wir{\'e}n, ``Compounding in a swedish blog corpus,''
  \emph{Computer mediated discourse across language. Stockholm: Stockholm
  University}, pp. 45--63, 2013.

\bibitem{kim:2016}
Y.~Kim, Y.~Jernite, D.~Sontag, and A.~M. Rush, ``Character-aware neural
  language models,'' in \emph{Proceedings of the Thirtieth AAAI Conference on
  Artificial Intelligence}.\hskip 1em plus 0.5em minus 0.4em\relax AAAI Press,
  2016, pp. 2741--2749.

\bibitem{KneserNey1995}
R.~Kneser and H.~Ney, ``Improved backing-off for m-gram language modeling,'' in
  \emph{1995 International Conference on Acoustics, Speech, and Signal
  Processing}, vol.~1, May 1995, pp. 181--184 vol.1.

\bibitem{ChenGoodman1999}
\BIBentryALTinterwordspacing
S.~F. Chen and J.~Goodman, ``An empirical study of smoothing techniques for
  language modeling,'' \emph{Computer Speech {\&} Language}, vol.~13, no.~4,
  pp. 359--393, 1999. [Online]. Available:
  \url{http://dx.doi.org/10.1006/csla.1999.0128}
\BIBentrySTDinterwordspacing

\bibitem{GelmanHill2006}
A.~Gelman and J.~Hill, \emph{Data Analysis Using Regression and
  Multilevel/Hierarchical Models}.\hskip 1em plus 0.5em minus 0.4em\relax
  Cambridge Univ Press, 2006.

\bibitem{Howell2010}
D.~C. Howell, \emph{Statistical Methods for Psychology}.\hskip 1em plus 0.5em
  minus 0.4em\relax Wadsworth, 2010.

\bibitem{r2014}
{R Core Team}, ``R: {A} {Language} and {Environment} for {Statistical}
  {Computing},'' Vienna, Austria, 2014.

\bibitem{kuznetsova2014}
A.~Kuznetsova, P.~B. Brockhoff, and R.~H. Bojesen~Christensen, ``{lmerTest}:
  {Tests} for random and fixed effects for linear mixed effect models (lmer
  objects of lme4 package).'' 2014, r package version 2.0-6.

\end{thebibliography}

\end{document}